\documentclass[a4paper]{doublecol-new}
\usepackage[latin9]{inputenc}
\usepackage{rotating}
\usepackage{float}
\usepackage{booktabs}
\usepackage{multirow}
\usepackage{amsmath}
\usepackage{amssymb}
\usepackage{graphicx}
\usepackage[authoryear]{natbib}

\makeatletter

\providecommand{\tabularnewline}{\\}
\floatstyle{ruled}
\newfloat{algorithm}{tbp}{loa}
\providecommand{\algorithmname}{Algorithm}
\floatname{algorithm}{\protect\algorithmname}

\@ifundefined{date}{}{\date{}}
\usepackage{natbib,stfloats}
\usepackage{mathrsfs}
\def\newblock{\hskip .11em plus .33em minus .07em}
\usepackage{algorithm}
\usepackage{algorithmic}
\floatname{algorithm}{Algorithm}

\@ifundefined{showcaptionsetup}{}{%
 \PassOptionsToPackage{caption=false}{subfig}}
\usepackage{subfig}
\makeatother

\begin{document}

\begin{abstract}
Auto-encoders are often used as building blocks of deep network classifier
to learn feature extractors, but task-irrelevant information in the
input data may lead to bad extractors and result in poor generalization
performance of the network.  In this paper,via dropping the task-irrelevant
input variables the performance of auto-encoders can be obviously
improved .Specifically, an importance-based variable selection method
is proposed to aim at finding the task-irrelevant input variables
and dropping them.It firstly estimates importance of each variable,and
then drops the variables with importance value lower than a threshold.
In order to obtain better performance, the method can be employed
for each layer of stacked auto-encoders. Experimental results show
that when combined with our method the stacked denoising auto-encoders
achieves significantly improved performance on three challenging datasets.
\end{abstract}

\setcounter{page}{1}
\LRH{Training Auto-encoder Effectively via Eliminating Task-irrelevant Input Variables}
\RRH{H. Shen*,D.Li,H. Wu,Z.Zang*}
\authorA{Hui~Shen*,~Dehua~Li~and~Hong~Wu~}
\affA{School of Automation \\Huazhong University of Science \& Technology \\1037 Luoyu Road, Wuhan, Hubei, 430074 China \\e-mail: shenhui@hust.edu.cn  *Corresponding author; \\lidehua1946@sina.com; \\wolfe\_wu@hust.edu.cn}

\authorB{Zhaoxiang~Zang*~}
\affB{Hubei Key Laboratory of Intelligent Vision Based Monitoring for Hydroelectric Engineering\\China Three Gorges University\\Yichang Hubei, 443002, China\\e-mail: zxzang@gmail.com  *Corresponding author }

\KEYWORD{feature learning; deep learning; neural network; auto-encoder; stacked auto-encoders; variable selection; feature selection; unsupervised training}

\begin{bio} Hui Shen is currently a PhD student in School of Automation, Huazhong University of Science and Technology,China. She obtained her bachelor's in Wuhan Polytechnic University, China.And got her master's in Huazhong University of Science and Technology, China. Her main fields of interest are neural network,deep learning and machine learning.

DeHua Li is a professor in School of Automation, Huazhong University of Science and Technology,China.He got his bachelor's from Wuhan University in 1970. And he spent one year as senior visiting scholar in AI department of University of Edinburgh, UK. His research interests including AI, neotic science, and machine learning.

Hong Wu obtained his bachelor's in computer science and technology from Wuhan University,China. And got his master's in Huazhong University of Science and Technology,China. Now he is a PhD student of the School of Automation, Huazhong University of Science and Technology, China. His main research fields are neural network, deep learning and machine learning.

Zhaoxiang Zang is an associate professor in College of Computer and Information Technology, China Three Gorges University, Yichang Hubei, China, and he is member of Hubei Key Laboratory of Intelligent Vision Based Monitoring for Hydroelectric Engineering, China Three Gorges University, China. He obtained his master's and PhD from Huazhong University of Science and Technology,China.His research interests mainly in the fields of machine learning,computational intelligence and computer game intelligence.
\end{bio}

\title{Training Auto-encoders Effectively via Eliminating Task-irrelevant
Input Variables}

\maketitle

\section{Introduction\label{sec:Introduction}}

Neural networks are widely applied in various fields, such as oil
exploration in \citet{Liu2011}, speech recognition in \citet{arisoy2015bidirectional},
temperature control in \citet{Admuthe2015}, and so on. As a kind
of neural network, deep neural network has become an increasingly
popular research field. And training auto-encoders to learn useful
feature extractors to inisitalize a deep neural network is a widely
used approach. An auto-encoder is comprised by an encoder and a decoder.
Given an input example, the encoder, which consists of a group of
feature extractors, produces features that constitute an abstract
representation or code of the example, while the decoder reconstructs
the example from the code. Training an auto-encoder is to minimize
the difference between the input example and its reconstruction. Details
of auto-encoders and their applications in deep learning can be found
in \citet{hinton2006fast,raey,ciresan2012multi,bengio2013representation}.
Auto-encoders are usually implemented with neural networks, but over-complete
(higher dimensional hidden layer than the input layer) and unconstrained
auto-encoders may learn identical mapping, which results in useless
features. Therefore, many regularized auto-encoders were proposed
to learn good feature extractors, such as sparse auto-encoders \citep{lee2008sparse,boureau2008sparse,ng2011sparse},
contractive auto-encoders (CAEs) \citep{rifai2011manifold,rifai2011higher,Rifai2011},
denoising auto-encoders (DAEs) \citep{vincent2008extracting,Vincent2010},
marginalized denoising auto-encoders (mDAEs) \citep{Chen2012,chen2014marginalized},
and so on.

Because of the unsupervised training, auto-encoders attempts to capture
everything in input data, including task-irrelevant information if
there exits\citep{Vincent2010}. However, learning task-irrelevant
information may waste the computation resources and the capability
of networks, and is easy to cause overfitting. Therefore, eliminating
task-irrelevant information becomes one of the ways to obtain better
performance and reduce computation cost. Bengio et al. explored supervised
pre-training (\citealp{Bengio2007Greedy}), and they concluded that
partially supervised pre-training (alternately perform supervised
and unsupervised training of an auto-encoder) can lead auto-encoders
to learn better feature extractors when much task-irrelevant information
is contained in training data. Shon et al. proposed point-wise gated
Boltzmann machines (PGBM) \citep{Kihyuk2013Learning}. They used a
hidden layer containing two-group units to model the foreground and
background respectively, then took the foreground group to extract
task-relevant features. PGMB achieves the state-of-the-art performance
on several benchmark datasets. Inspired by human attention, Wang et
al. proposed the attentional neural network (aNN) \citep{Wang2014}.
They used a segmentation module to iteratively segment foreground
from noisy input via a feedback loop, and employed a deep network
on the foreground for classification. aNN also achieves the state-of-the-art
performance on one benchmark dataset.

In this work, we try to drop the task-irrelevant variables by performing
variable selection on input of auto-encoders and then execute unsupervised
learning on the remaining variables. We introduce an importance-based
variable selection method that evaluate importance of each variable
and drop the variables with low importance.For obtaining better classification
performance, the method is employed for each layer of stacked auto-encoders,
not only on the raw input for the first layer, which is different
from methods mentioned above. The experiment results show that it
helps stacked DAE (SDAE) achieve significantly improved performance
on three challenging benchmark datasets.

The rest of this paper is organized as following. Preliminaries about
DAE and SDAE are provided in section \ref{sec:Preliminaries}. The
proposed variable selection method is described in detail in section
\ref{sec:Methods}. Experiments and results are reported in section
\ref{sec:Experiments-and-Results}, and conclusions are made in section
\ref{sec:Conclusion}. We will use following notations through out
the paper. $\boldsymbol{\chi}$ is a training dataset. Each element
of $\boldsymbol{\chi}$ contains a input example $\boldsymbol{x}$
and a target label $r$ such that $\boldsymbol{x}\in[0,1]^{M}$ (or
$\mathbb{R}^{M}$) and $r\in\left\{ 1,\ldots,K\right\} $. For a vector
$\boldsymbol{x}$, $x_{d}$ is its $d$-th component. For the sake
of simplicity, we use $\boldsymbol{\chi}$ to denote the training
dataset for an auto-encoder, no matter where the auto-encoder is located
in a stacked auto-encoders.

\section{Preliminaries\label{sec:Preliminaries}}

Auto-encoders(AEs) are often employed as building blocks of deep networks.
An AE is a neural network that composed of three layers,an input layer,
a hidden layer and an output layer. It tries to recover the input
data $\boldsymbol{x}$ from the hidden representation $\boldsymbol{h}$
at the output layer. The motivation is that, if the input data can
be reconstructed well enough, then it can be said that the hidden
feature is a discription of the input data. An AE is also can be seen
as a network that consists of an encoder and a decoder. Processing
from input layer to hidden layer can be seen as an encoder, while
from hidden layer to output layer a decoder.

During training,it firstly maps the the input data $\boldsymbol{x}$
to hidden representation $\boldsymbol{h}$ by the encoder:
\[
h_{q}=f_{q}\left(\boldsymbol{x}\right)=s_{f}\left(\boldsymbol{w}_{q}^{T}\boldsymbol{x}+b_{q}\right),
\]
where $\boldsymbol{w}_{q}^{T}$ is a weight vector, $b_{q}$ is a
bias and $s_{f}$ is the activation function of the encoder, typically
the sigmoid function$s_{f}\left(z\right)=1/\left(1+e^{-z}\right)$.
Then reconstruct input $y_{d}$ from the hidden representation $\boldsymbol{h}$
through the decoder:
\[
y_{d}=g_{d}\left(\boldsymbol{h}\right)=s_{g}\left(\boldsymbol{w}_{d}^{T}\boldsymbol{h}+c_{d}\right),
\]
where $c_{d}$ is a bias and $s_{g}$ is the activation function of
the decoder, which can be sigmoid function for binary input or an
identity for continuous input.

Explicitly, the reconstruction $\boldsymbol{y}$ should approach to
original input $\boldsymbol{x}$ as much as possible, this can be
measured by the reconstruction error $L\left(\boldsymbol{x},\boldsymbol{y}\right)$
which is typically computed via squared error or cross-entropy. Which
one to be chosen depends on the activation function of the decoder. 

If $s_{g}$ is an identity, i.e.$s_{g}\left(z\right)=z$, then
\[
L\left(\boldsymbol{x},\boldsymbol{y}\right)=||\boldsymbol{x}-\boldsymbol{y}||^{2}.
\]

If $s_{g}$ is simoid function, i.e.$s_{g}\left(z\right)=1/\left(1+e^{-z}\right)$,
then
\begin{equation}
L\left(\boldsymbol{x},\boldsymbol{y}\right)=-\sum_{d=1}^{d=M}x_{d}log\left(y_{d}\right)+\left(1-x_{d}\right)log\left(1-y_{d}\right),\label{eq:reconstruction_error}
\end{equation}
where $y_{d}$ is depend on the model parameters $w_{q}$,$w_{d}$,$b_{q}$
and $c_{d}$when input data $x$ is given. 

Then the auto-encoder is trained to learn model parameters that minimize
the reconstruction error $L\left(\boldsymbol{x},\boldsymbol{y}\right)$.
Gradient descent can be utilized to optimized the error function during
training. When the training is completed, the output layer with the
weights of hidden to output are dropped and the learned representation
feature holds in the hidden layer, which can be used for classification
or used as the input of an other autoencoder to learn more abstract
feature.

However, an AE may learn identity, which leads to obtain no useful
featrues. In order to prevent this situation, AEs often utilize the
configuration called ``bottleneck'' of which the quantity of hidden
units lower than input units. An other approch is adding regular terms
on the objective function to constrain the weights. Otherwise, using
disturbed input data for training is also an effective means, like
denoising autoencoder.

Denoising auto-encoder(DAE) is a variant of standard auto-encoder.
It attempts to reconstruct the input $\boldsymbol{x}$ from the encoded
representation $\boldsymbol{h}$ of noisy input $\tilde{\boldsymbol{x}}$
via a decoder. By disturbing the input $\boldsymbol{x}$, denoising
auto-encoder tries to learn robust features that can successfully
recover the perturbed values to reconstruct the original input data.
If a DAE can recover the original input data from the code of corruped
input data, it can be said that the DAE has leanred robust and stable
features. 

Stacked denoising auto-encoder(SDAE) is formed by stacking multiple
single-layer DAEs for learning more abstract representations. SDAE
can be used to effectively pre-train deep networks. In the process
of pre-training by SDAE, the hidden features learned by lower-layer
DAE are used as inputs for training next (upper-layer) DAE, and the
encoders of DAEs are used to initialize weights in the deep network.
See \citet{Vincent2010} for details.

\section{Importance-based Variable Selection \label{sec:Methods}}

According to equation (\ref{eq:reconstruction_error}), AEs belong
to unsupervised learning without considering the label information.
The hidden representation of an autoecoder is a description of the
whole input data. AEs do not identify useful or unuseful information
for classification. they attempt to capture all the information of
the input, not only task-relevant information, but also task-irrelevant
information if there exists. However, learning task-irrelevant information
may waste computation resources and even cause over-fitting. Therefore,
task-irrelevant information contained in input data should be reduced
or eliminated for obtaining better performance.

To address this issue, an importance-based variable selection method
is proposed to find the task-irrelevant variables and drop them. Briefly,
the method is to evaluate the importance of each variable to classification
and drop the variables with importance lower than a threshold. We
exploit the sensitivity of the discriminant hyperplane to a variable
to evaluate the importance of the variable. we argue that, the variables
with higher sensitivity are more important for classification, these
variables also possess higher importance value and should be reserved,
while those variables with low importance(lower than a threshold)
should be dropped. The details are described as follows.

We employ a trained Multinominal Logistic Regression(MLR) model as
a pre-classifier to help us determine the importance of each input
variable to classification. Multinominal Logistic Regression (MLR)
is a simple log-linear classifier, and can be easily analyzed. Given
an example $\boldsymbol{x}$, the MLR computes the posterior probability
of each hypothesis via a softmax function, and takes the one with
biggest posterior probability as prediction. see \citet{bishop2006pattern,hosmer2013applied}
for MLR in detail. We briefly introduce the softmax function, from
which the importance notation can be deduced.

The softmax function can be written as 
\[
\sigma_{i}\left(\boldsymbol{x}\right)=\exp\left(\boldsymbol{w}_{i}^{T}\boldsymbol{x}+b_{i}\right)\slash\left[\sum_{c=1}^{K}\exp\left(\boldsymbol{w}_{c}^{T}\boldsymbol{x}+b_{c}\right)\right]
\]
 where $\boldsymbol{w}_{i}$ and $b_{i}$ are parameters, and $i\in\left\{ 1,\ldots,K\right\} $
is the index of class. The predicted class of $\boldsymbol{x}$ is
obtained by $y=\mathop{\arg\max}_{i}\sigma_{i}\left(\boldsymbol{x}\right)$. 

The softmax function computes the estimated probability of the class
label for a given input $\boldsymbol{x}$. Now we consider any two
classes of class i and class j. We suppose that, the discriminant
hyperplane between class $i$ and class $j$ is consisted of the points
that with equivalent estimated probabilities of the two labels. Let
\[
\frac{\exp\left(\boldsymbol{w}_{i}^{T}\boldsymbol{x}+b_{i}\right)}{\sum_{c=1}^{K}\exp\left(\boldsymbol{w}_{c}^{T}\boldsymbol{x}+b_{c}\right)}=\frac{\exp\left(\boldsymbol{w}_{j}^{T}\boldsymbol{x}+b_{j}\right)}{\sum_{c=1}^{K}\exp\left(\boldsymbol{w}_{c}^{T}\boldsymbol{x}+b_{c}\right)},
\]
then we obtain the discriminant hyperplane:
\[
\left(\boldsymbol{w}_{i}-\boldsymbol{w}_{j}\right)^{T}\boldsymbol{x}+\left(b_{i}-b_{j}\right)=0
\]

After normalization, discriminant function between class $i$ and
class $j$ can be written as 
\[
f_{i,j}\left(\boldsymbol{x}\right)=\left[\left(\boldsymbol{w}_{i}-\boldsymbol{w}_{j}\right)^{T}\boldsymbol{x}+\left(b_{i}-b_{j}\right)\right]\slash\|\boldsymbol{w}_{i}-\boldsymbol{w}_{j}\|_{2}
\]
In other words, all $\boldsymbol{x}$ that satisfy $f_{i,j}\left(\boldsymbol{x}\right)=0$
form a discriminant hyperplane between class $i$ and class $j$.
We denote the discriminant hyperplane as $\mathcal{H}_{i,j}$. The
unit normal vector of $\mathcal{H}_{i,j}$ can be written as

\begin{equation}
\boldsymbol{v}_{i,j}=\frac{\boldsymbol{w}_{i}-\boldsymbol{w}_{j}}{\|\boldsymbol{w}_{i}-\boldsymbol{w}_{j}\|_{2}}.\label{eq:normal-vector}
\end{equation}

\begin{algorithm*}[t]
\caption{\label{alg:IVS}Importance-based Variables Selection }

\algsetup{indent=2em}

\begin{algorithmic}[1]

\REQUIRE training dataset $\boldsymbol{\chi}=\left\{ \boldsymbol{x}^{\left(t\right)},\ r^{\left(t\right)}\right\} _{t=1}^{N}$,
importance threshold $c_{th}$

\ENSURE variables mask $\boldsymbol{\alpha}$

\STATE$\boldsymbol{\alpha}\leftarrow\left(\boldsymbol{1}\left(true\right)\right)_{d=1}^{M}$\label{initilize-a}

\LOOP

\STATE Train a new pre-classifier MLR with the masked training data
$\left\{ \boldsymbol{\alpha}\odot\boldsymbol{x}^{\left(t\right)},\ r^{\left(t\right)}\right\} _{t=1}^{N}$
. \label{train-MLR}

\IF{stop criterion}

\RETURN$\boldsymbol{\alpha}$

\ELSE

\STATE Update $\boldsymbol{\alpha}$ according to (\ref{eq:normal-vector}),
(\ref{eq:hyperplane-contribution}), (\ref{eq:task-contribution}),
and (\ref{eq:task-mask}). \label{update-a}

\ENDIF

\ENDLOOP

\end{algorithmic}
\end{algorithm*}

Define the \emph{sensitivity} of $f_{i,j}$ to an input variable as
$\lvert\frac{\partial f_{ij}}{\partial x_{d}}\rvert$, which reflects
the influence of the variable to $f_{i,j}$. Because $f_{i,j}$ is
a linear function, $\lvert\frac{\partial f_{ij}}{\partial x_{d}}\rvert=\lvert v_{i,j,d}\text{\ensuremath{\rvert}}$
where $v_{i,j,d}$ is the $d$-th component of $\boldsymbol{v}_{i,j}$.
By normalizing the sensitivities of $f_{i,j}$ with $\|\boldsymbol{v}_{i,j}\|_{\infty}$
(the infinity norm of $\boldsymbol{v}_{i,j}$), we can define the
\emph{importance} of the $d$-th variable \emph{to} $f_{i,j}$ as

\begin{equation}
s_{i,j,d}=\frac{\lvert v_{i,j,d}\text{\ensuremath{\rvert}}}{\|\boldsymbol{v}_{i,j}\|_{\infty}}=\frac{\lvert v_{i,j,d}\text{\ensuremath{\rvert}}}{\max_{k}\lvert v_{i,j,k}\rvert}.\label{eq:hyperplane-contribution}
\end{equation}

We argue that if a variable has low importances to all the discriminant
functions then it can be identified as task-irrelevant variable. On
the other hand, a variable is identified as task-relevant variable
if it has unignorable importance to any $f_{i,j}$. Therefore, we
define the \emph{importance} of the $d$-th variable \emph{to the
classification} as

\begin{equation}
c_{d}=\max_{i,j\neq i}s_{i,j,d}.\label{eq:task-contribution}
\end{equation}
$c_{d}$ is the maximum of importances of the $d$-th variable across
all discriminant functions.

Consequently, task-irrelevant variables, each of which has low importance
(below a threshold) to classification, can be discarded in the unsupervised
training of auto-encoder. In order to facilitate computation, we use
a variable mask $\boldsymbol{\alpha}$ to represent the binary task-relevances
of input variables. $\boldsymbol{\alpha}$ is defined as 

\begin{equation}
\boldsymbol{\alpha}=\left(\boldsymbol{1}\left(c_{d}\geqslant c_{th}\right)\right)_{d=1}^{M},\label{eq:task-mask}
\end{equation}
where $c_{th}$ is a importance threshold and $\boldsymbol{1}\left(\cdot\right)$
is the indicator function so that it takes 1 if the condition in the
brackets is true and 0 otherwise. Mask components corresponding to
task-irrelevant variables will take 0.

In practice, since there might be cross-correlation between variables
in input variable set, it is not easy to find out all task-irrelevant
variables through training a MLR with full input variable set. A iterative
method can be employed to find out task-irrelevant variables gradually,
and in each iteration a new pre-classifier MLR is trained to dropping
a few variables from input variable set.

Algorithm \ref{alg:IVS} is called Importance-based Variable Selection
(IVS), and describes how to find task-irrelevant variables in detail.
In line \ref{initilize-a}, variable mask $\boldsymbol{\alpha}$ is
initialized by assigning 1 to each component, which means all input
variables will be used in the first MLR training. For each iteration,
a new pre-classifier MLR is trained with masked training data in line
\ref{train-MLR}, where $\odot$ means component-wise multiplication
or Haddamard product. In order to prevent the MLR training from overfitting,
model selection can be done by using a validation dataset to early
stop the training in line \ref{train-MLR}. The variable mask $\boldsymbol{\alpha}$
is updated in line \ref{update-a} based on the well-trained MLR.
This iterative procedure will stop under conditions such as exceeding
maximum iterations, no more task-irrelevant variables found, no better
classification performance obtained on validation set and so on. Once
the variable mask $\boldsymbol{\alpha}$ is obtained, task-irrelevant
variables indicated by $\boldsymbol{\alpha}$ will be dropped in the
following unsupervised training for auto-encoders.

Algorithm \ref{alg:IVS} can be employed for each higher layer of
stacked auto-encoders, therefore complex task-irrelevant information
not eliminated on low level can be removed gradually on higher layers.
From a model selection point of view, eliminating task-irrelevant
variables can reduce complexity of networks therefore obtain better
performance.

\section{Experiments and Results\label{sec:Experiments-and-Results}}

\begin{figure*}
\begin{centering}
\includegraphics[scale=0.32]{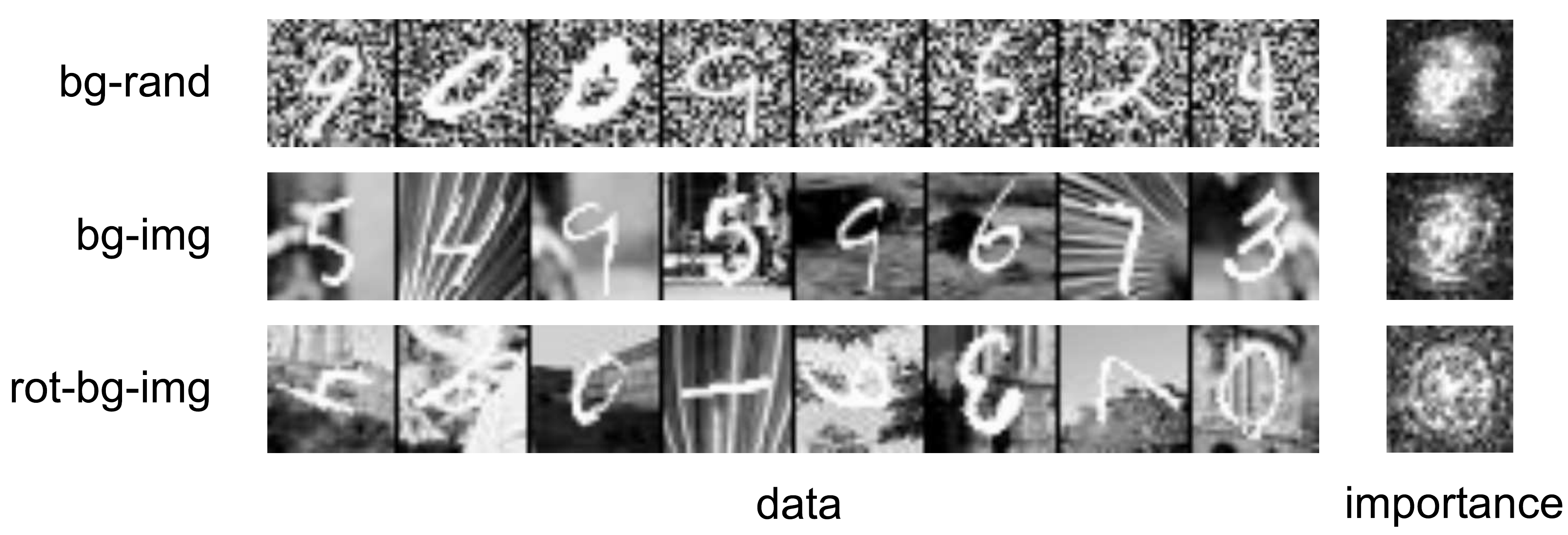}
\par\end{centering}

\caption{Inputs examples(left) and visualization of importances of variables
(right). The importances are obtained from MLRs trained with full
variable sets. The light of point denotes the importance of corresponding
variable to classification. White stands for 1 and black for 0.}

\label{fig:importance}
\end{figure*}

\begin{figure*}[t]
\begin{centering}
\subfloat[Task-irrelevant pattern]{\begin{centering}
\includegraphics[scale=0.32]{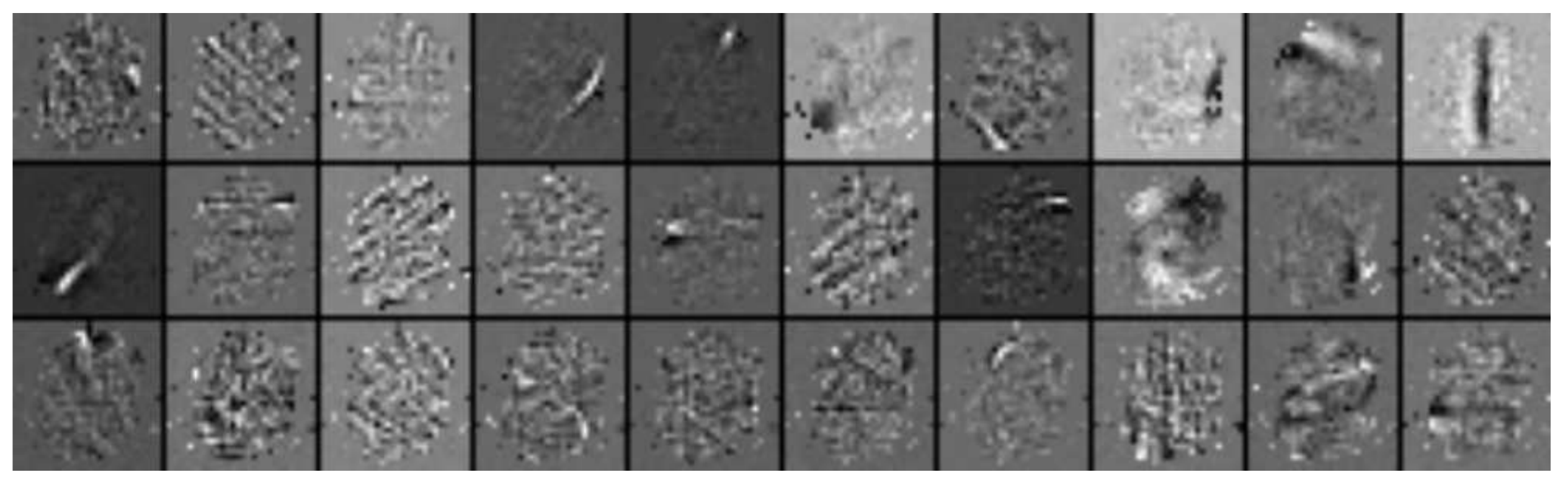}
\par\end{centering}

}\subfloat[Task-relevant pattern]{\begin{centering}
\includegraphics[scale=0.32]{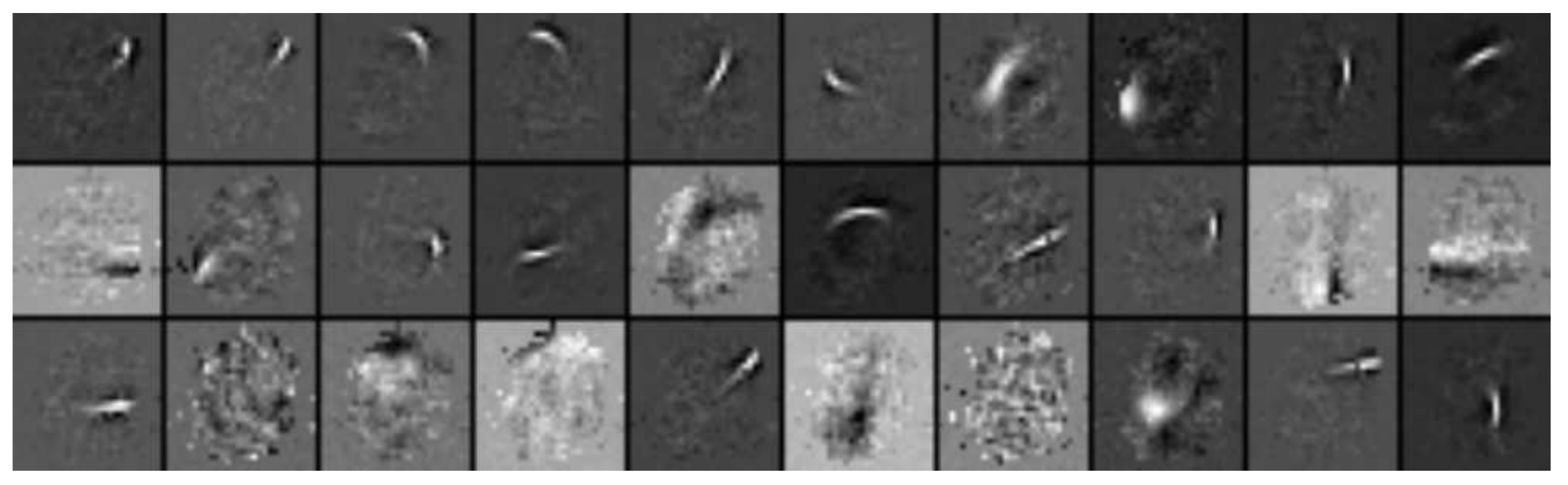}
\par\end{centering}

}
\par\end{centering}

\caption{Visualization of learned patterns on the first hidden layer. Importance
threshold is 0.3. The data set is bg-img.}
\label{fig:features}
\end{figure*}

\begin{figure*}[t]
\begin{centering}
\includegraphics[scale=0.6]{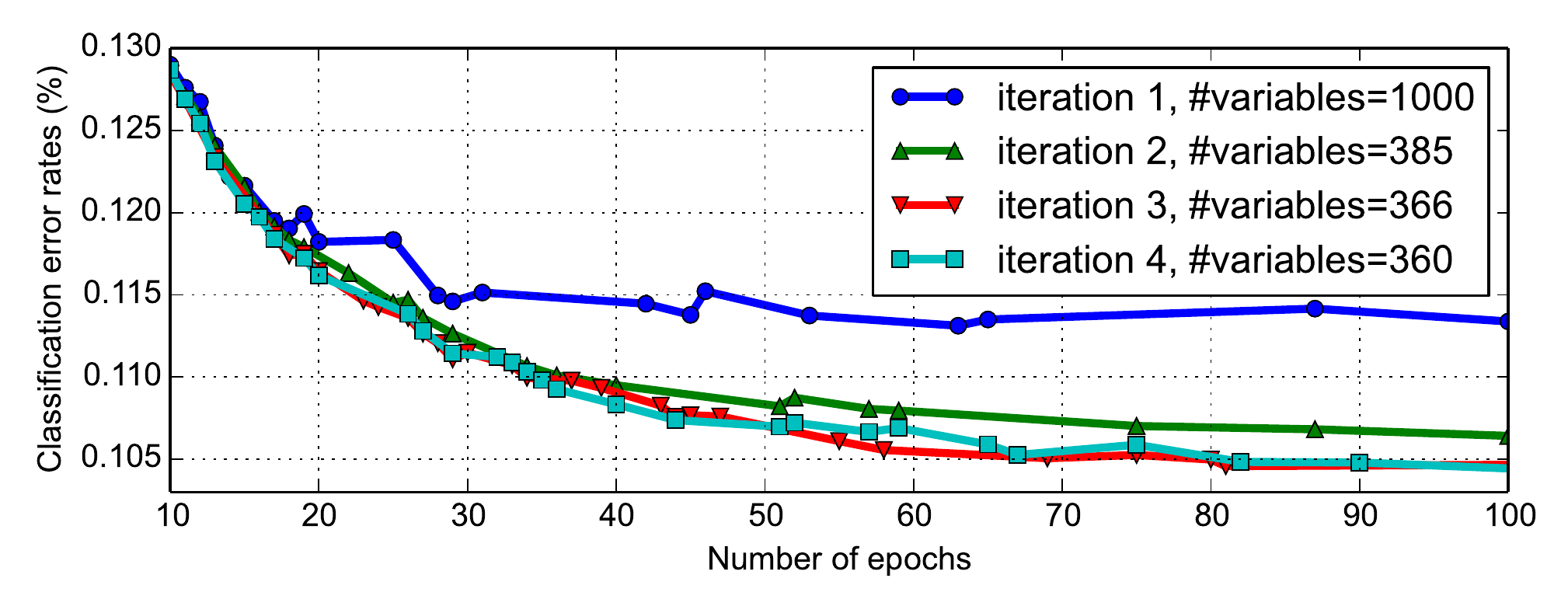}
\par\end{centering}

\caption{Test classification error rates (in \%) on bg-rand at different training
epochs of MLRs. Each MLR is the pre-classifier in a iteration of Algorithm
\ref{alg:IVS} that performs upon the first hidden layer of SDAE-IVS.
The importance threshold is 0.3, and the Gaussian noise standard deviation
in SDAE-IVS is 0.2.}

\label{fig:iteration_performance}
\end{figure*}

\begin{figure*}[t]
\begin{centering}
\includegraphics[scale=0.6]{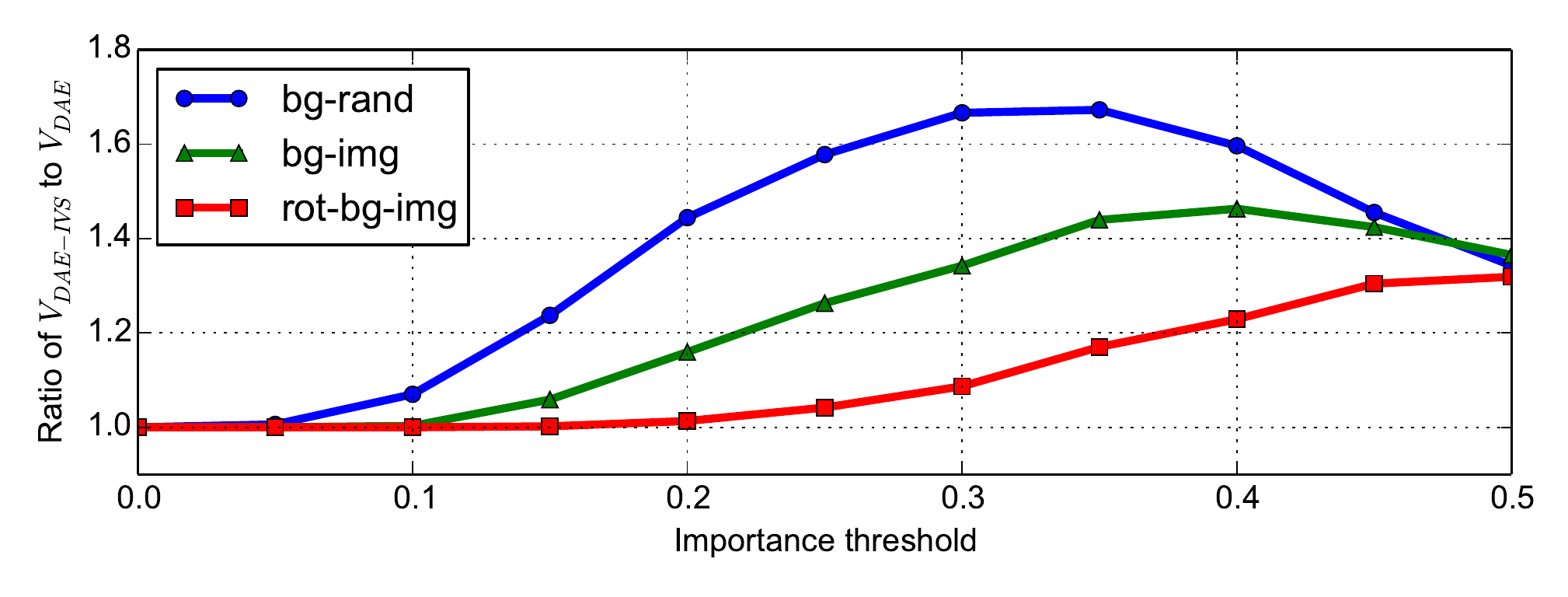}
\par\end{centering}

\caption{Ratio of $V_{DAE-IVS}$ to $V{}_{DAE}$ (see the text for explanation)
against importance threshold on first hidden layer. }

\label{fig:number_ratio}
\end{figure*}

\begin{figure*}[t]
\noindent \begin{centering}
\subfloat[SDAE on bg-rand]{\includegraphics[scale=0.32]{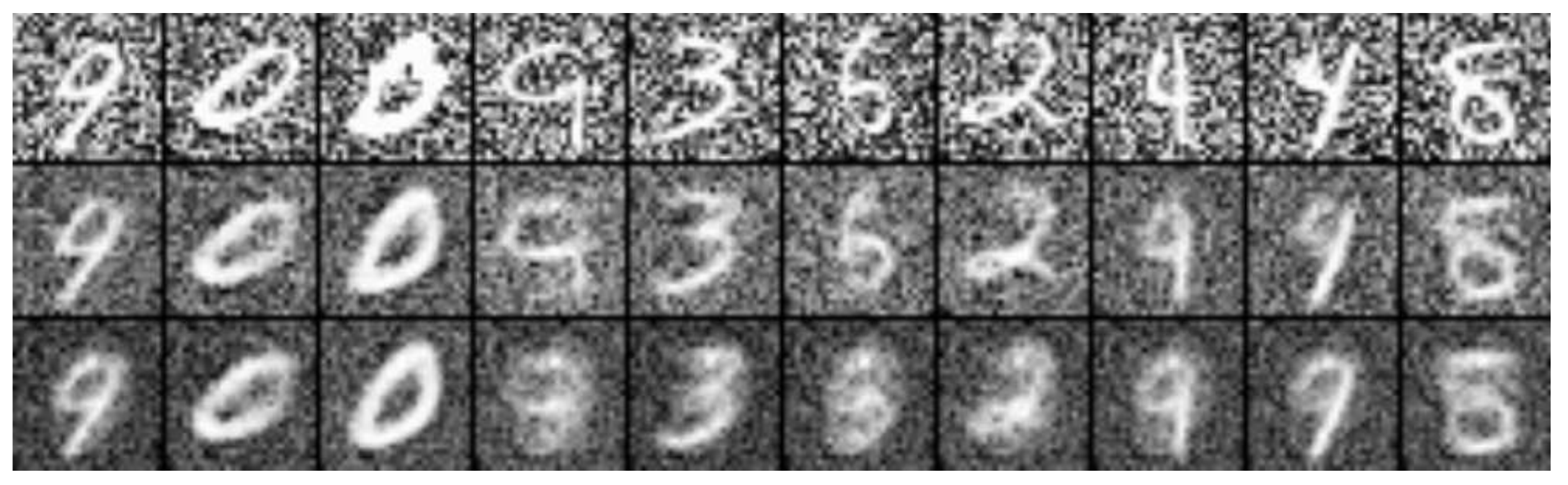}

}\ \ \subfloat[SDAE-IVS on bg-rand]{\includegraphics[scale=0.32]{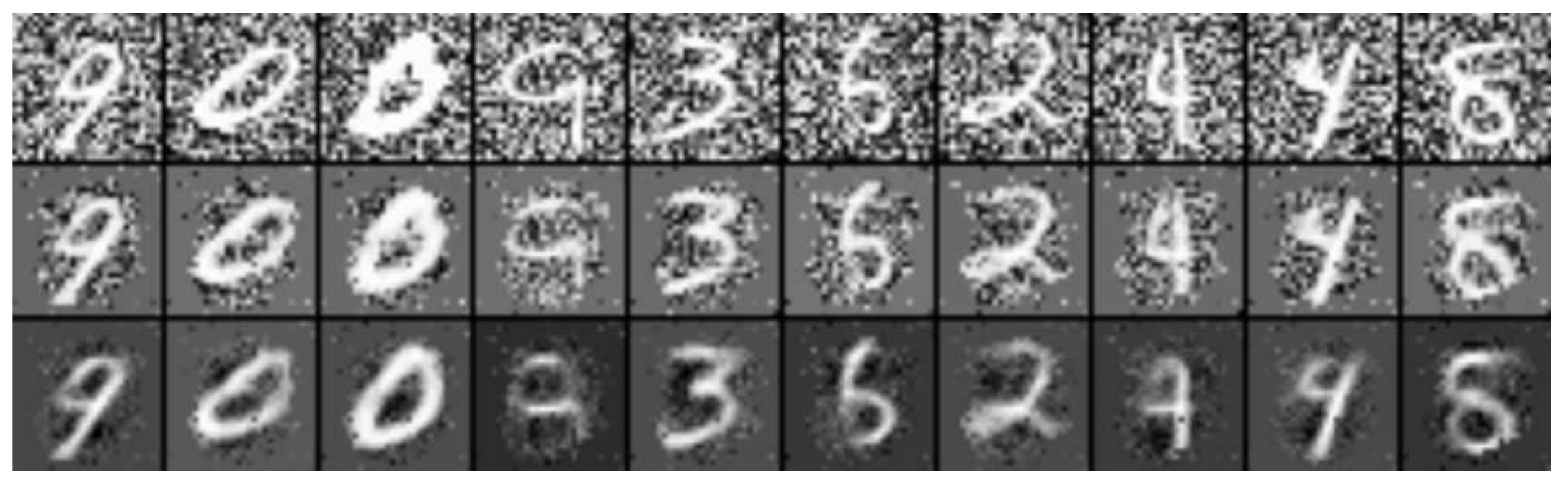}

}
\par\end{centering}

\begin{centering}
\subfloat[SDAE on bg-img]{\includegraphics[scale=0.32]{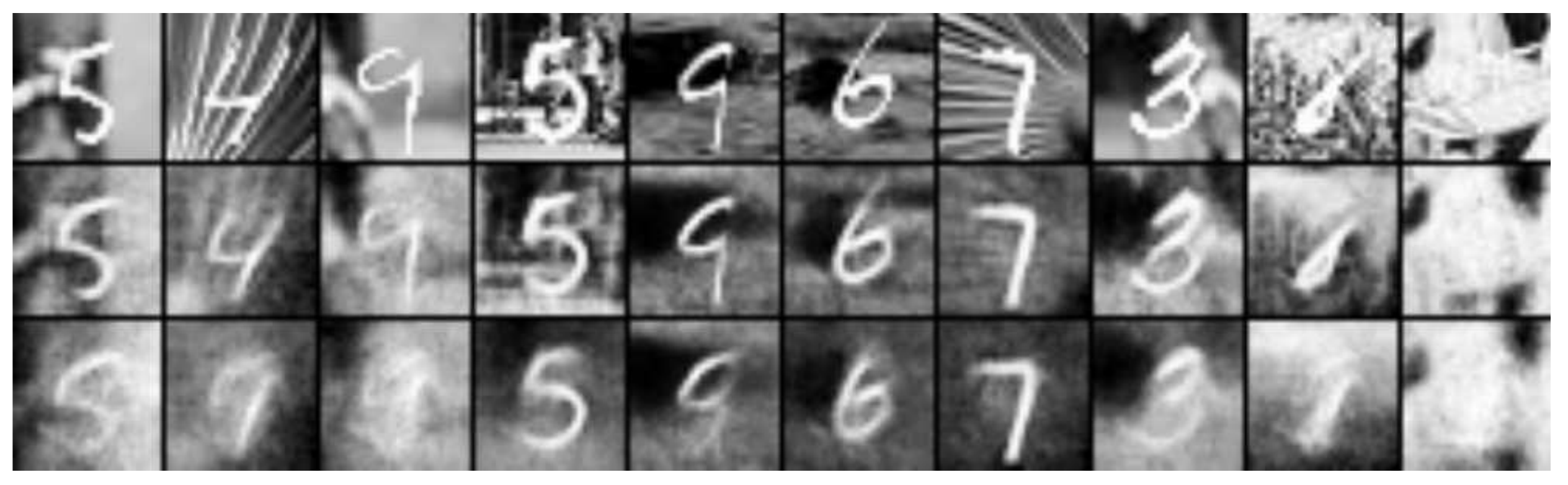}

}\ \ \subfloat[SDAE-IVS on bg-img]{\includegraphics[scale=0.32]{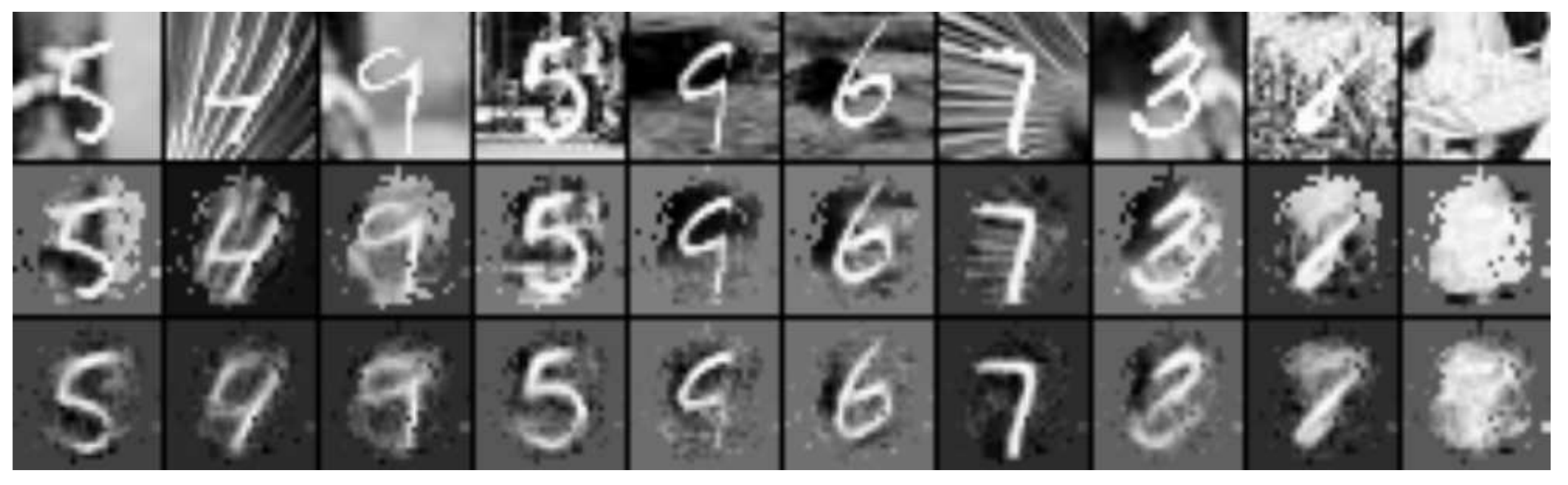}}
\par\end{centering}

\caption{Reconstructions produced by SDAE and SDAE-IVS. The first row is origin
data, the second and the third rows are produced with 1 and 2 stacked
layers respectively.}

\label{fig:recosntruction}
\end{figure*}

\begin{table*}[t]
\caption{Test classification error rates (in \%) produced by different depth
of SDAE and SDAE-IVS. A 95\% confidence interval for each result is
also given. Best performer is in bold.}
\label{tab:classification}

\centering{}%
\begin{tabular}{clccc}
\toprule 
\multicolumn{2}{c}{\textbf{Methods}} & bg-rand & bg-img & rot-bg-img\tabularnewline
\midrule
\midrule 
\addlinespace[0.005\textheight]
\multirow{2}{*}{\begin{turn}{90}
1 Layer
\end{turn}} & \textbf{SDAE-1} & $11.85\pm0.28$ & $14.82\pm0.31$ & $46.24\pm0.44$\tabularnewline\addlinespace[0.005\textheight]
\cmidrule{2-5} 
\addlinespace[0.005\textheight]
 & \textbf{SDAE-IVS-1} & $\mathbf{9.92}\pm0.26$ & $\mathbf{13.58}\pm0.30$ & $\mathbf{44.72}\pm0.43$\tabularnewline\addlinespace[0.005\textheight]
\midrule 
\addlinespace[0.005\textheight]
\multirow{2}{*}{\begin{turn}{90}
2 Layers
\end{turn}} & \textbf{SDAE-2} & $10.00\pm0.26$ & $13.41\pm0.30$ & $41.05\pm0.43$\tabularnewline\addlinespace[0.005\textheight]
\cmidrule{2-5} 
\addlinespace[0.005\textheight]
 & \textbf{SDAE-IVS-2} & $\mathbf{7.58}\pm0.23$ & $\mathbf{12.31}\pm0.28$ & $\mathbf{39.35}\pm0.43$\tabularnewline\addlinespace[0.005\textheight]
\midrule 
\addlinespace[0.005\textheight]
\multirow{2}{*}{\begin{turn}{90}
3 Layers
\end{turn}} & \textbf{SDAE-3} & $9.50\pm0.26$ & $13.06\pm0.29$ & $39.88\pm0.43$\tabularnewline\addlinespace[0.005\textheight]
\cmidrule{2-5} 
\addlinespace[0.005\textheight]
 & \textbf{SDAE-IVS-3} & $\mathbf{7.05}\pm0.23$ & $\mathbf{11.21}\pm0.27$ & $\mathbf{37.53}\pm0.43$\tabularnewline\addlinespace[0.005\textheight]
\bottomrule
\end{tabular}
\end{table*}

In our experiments, we combined the proposed method with DAE \citet{Vincent2010},
which is called DAE-IVS, and compared performances produced by stacked
DAE-IVS (SDAE-IVS), stacked DAE (SDAE) \citet{Vincent2010}, PGBM
\citet{Kihyuk2013Learning}, and aNN \citet{Wang2014} . The baseline
was SDAE trained with standard training strategy described in \citet{Vincent2010}.
We tested different depth of SDAE and SDAE-IVS from 1 layer to 3 layers.
Each auto-encoder had 1000 hidden units and used tied weights. Both
encoder and decoder used sigmoid function and took cross-entropy loss
as reconstruction error. The outputs of feature extractors learned
by a layer were used as input variables of upper layer. Multinominal
Logistic Regression layer was added on both top of SDAE and SDAE-IVS
to perform supervised fine-tuning. All training processes used stochastic
gradient descent for parameter learning. Our datasets were several
variants of MNIST dataset for recognizing images of handwritten digits
\citet{larochelle2007empirical}, including MNIST with random background(bg-rand)
or with image background (bg-img) and the combination of rotated digits
with image background (rot-bg-img). Each dataset was split into three
subsets: a training set (10000 examples) for pre-training and fine-tuning
the parameters, a validation set (2000 examples) for model selection
and a testing set (50000 examples) on which the classification performance
were reported. The hyper-parameters were chosen on the validation
set, including the learning rate for pre-classification in Algorithm
\ref{alg:IVS} (candidate set {[}0.01, 0.02, 0.05, 0.1{]}), the importance
threshold (candidate set {[}0.2, 0.25, 0.3, 0.35, 0.4, 0.45, 0.5{]}),
learning rate for pre-training and fine-tuning (candidate set {[}0.01,
0.05, 0.1, 0.2{]}), Gaussian noise standard deviation in SDAE and
SDAE-IVS (candidate set {[}0.1, 0.15, 0.2, 0.25, 0.3, 0.35, 0.4{]}),
and pre-training epochs (candidate set {[}60, 120, 180, 240, 300{]}).
The loop in Algorithm \ref{alg:IVS} was stopped when no more task-irrelevant
variables found and no better classification performance obtained
on validation set.

\subsection{Effect of importance-based variable selection}

During Applying algorithm\ref{alg:IVS}, we can compute the importance
value of each variable on each dataset(equation (\ref{eq:normal-vector}),
(\ref{eq:hyperplane-contribution}) and (\ref{eq:task-contribution})).
Examples and visualization of the importance of variable for each
dataset are shown in Figure \ref{fig:importance}. In importance
image, the lighter and the darker areas correspond to the variables
with higher importance and lower importance respectively. It can be
seen that variables with high importance concentrate on the central
area where digits mainly occupy. Different from other datasets, the
high-importance area of rot-bg-img looks like a disc because of the
rotation of digits. We take the variables with importance value
higher than a threshold as the task-relevant variable, otherwise as
the task-irrelevant ones (equation (\ref{eq:task-mask})). Visualizations
of the task-irrelevant and the task-relevant patterns (weights of
feature extractors learned by the first layer of SDAE-IVS) are showed
in Figure \ref{fig:features}, where the threshold is set by experience.
Although there exists misclassification, most task-irrelevant patterns
describe the background image, and most task-relevant patterns describe
the foreground digit. The algorithm \ref{alg:IVS} is an iterate
process, and task-irrelevant variables are dropped in each iteration.
This can indirectly improve the signal to noise ratio related to classification.
Figure \ref{fig:iteration_performance} shows that not only the number
of variables is decreased (to 36\%), but also the classification performance
of the pre-classifier is improved. Similar results can also be found
on different layers of SDAE-IVS trained on different datasets. 

Let $V_{DAE-IVS}$ and $V{}_{DAE}$ be the number of task-relevant
feature extractor learned by the first layer of SDAE-IVS and SDAE
respectively. We used the the ratio of $V_{DAE-IVS}$ to $V_{DAE}$
to measure the effectiveness improvement of learning useful feature
extractors. As shown in Figure \ref{fig:number_ratio}, all these
curves are above 1, which means that DAE-IVS could learn more task-relevant
feature extractor than DAE.   With the importance threshold increasing,
many actual task-relevant feature extractors were also dropped, thus
these curves get lower. However, they are still above 1.  Note that
we do not use the curves to optimize the threshold of importance,
we just concentrate on illustrating that feature selection is beneficial
to training AE.For showing the effect of feature selection, we try
to reconstruct the lower-layer data through the decoders of AEs by
using just the task-relevant features in higher-layer. By observing
the reconstruction result, we can see whether our algorithm effectively
eliminates the task-irrelevant variables and preserves the task-relevant
ones. In Figure \ref{fig:recosntruction}, we show the reconstructions
of raw data produced by SDEA and SDAE-IVS with different depth on
different datasets. The reconstructions are clearer after dropping
task-irrelevant variables, and the background information is significantly
suppressed.

\subsection{Classification performance comparison}

In Table \ref{tab:classification}, we show the test classification
error rate of produced by SDAE and SDAE-IVS with different depth.
It can be seen that in each depth the performance produced by SDAE-IVS
significantly outperforms the performance of SDAE. These results suggest
that our method can effectively help auto-encoders learn more and
better task-relevant feature extractor so as to get better task performance.

\section{Conclusion\label{sec:Conclusion}}

Auto-encoders attempt to capture as much as possible of information
in the input data and have to expend part of its capacity to learn
task-irrelevant information if there exists. More importantly, task-irrelevant
information may lead the eventual classification to overfitting resulting
in bad performance.

The proposed method is a simple and effective variable selection method
to deal with this problem. Through several rounds of variable selection,
the remaining input variables are fed into an auto-encoder to learn
feature extractors. Because this method is employed for each layer
of stacked auto-encoders, it not only eliminates task-irrelevant information,
but also prunes the deep network in a certain degree so as to efficiently
control the model complexity to obtain better performance. Experimental
results show that the method can efficiently drop task-irrelevant
variables and helps the auto-encoders learn more and better feature
extractor. It helps SDAE achieve significant improvements on classification
performances. 

In the future, we will explore some variable selection method that
deduced from non-linear classification models, expecting to help stacked
auto-encoders get better performance.

\section*{Acknowledgment }

This work was partially supported by the Doctoral Startup Foundation
of China Three Gorges University (Grant No. KJ2013B064), Natural Science
Foundation of Hubei (Grant Nos. 2015CFB336) and National Natural Science
Foundation of China (Grant No. 61502274). 

\bibliographystyle{apalike}

\end{document}